\documentclass{tx}

\usepackage{amsmath,amsfonts}
\usepackage{algorithmic}
\usepackage{algorithm}
\usepackage{array}
\usepackage[caption=false,font=normalsize,labelfont=sf,textfont=sf]{subfig}
\usepackage{textcomp}
\usepackage{stfloats}
\usepackage{url}
\usepackage{verbatim}
\usepackage{graphicx}
\usepackage{xcolor}
\usepackage{booktabs}
\usepackage{colortbl}
\usepackage{diagbox} 
\usepackage{multirow}
\usepackage{cite}
\usepackage{xspace}
\usepackage{pgfplots}

\usepackage{pifont} 
\usepackage{xcolor}

\usepackage{tikz}
\newcommand{\barpercent}[1]{%
  \begin{tikzpicture}[baseline=-0.5ex]
    \fill[blue!5!white] (0,0) rectangle (2.5,0.25); % 背景
    \fill[blue!35!white] (0,0) rectangle (2.5*#1/100,0.25); % 填充条
    \node[anchor=center, font=\scriptsize, text=black] at (1.25,0.125) {#1\%};
  \end{tikzpicture}%
}

\usepackage[dvipsnames]{xcolor}
\definecolor{metablue}{HTML}{0064E0}
\definecolor{metafg}{HTML}{1C2B33}
\definecolor{metabg}{HTML}{F1F4F7}

\usepackage{hyperref}
\hypersetup{
  colorlinks,
  linkcolor=metablue,
  citecolor=metablue,
  urlcolor=metablue
}

\newcommand{\name}{SeGPruner\xspace}
\newcommand{\imp}{Saliency-aware Token Selector\xspace}
\newcommand{\diversity}{Geometry-aware Token Diversifier\xspace}

\title{\name: Semantic–Geometric Visual Token Pruner for 3D Question Answering}
\author{
    Wenli Li,
    Kai Zhao,
    Haoran Jiang,
    Enquan Yang,
    Yi Su,
    Dan Zeng
}
\affiliation{
    Shanghai University, Shanghai, China
}
\email{
    \url{https://github.com/intcomp/SegPruner}
}

\begin{abstract}
Vision--language models (VLMs) have been widely adopted for 3D question answering (3D QA).
In typical pipelines, visual tokens extracted from multiple viewpoints are concatenated with language tokens and jointly processed by a large language model (LLM) for inference.
However, aggregating multi-view observations inevitably introduces severe token redundancy, leading to an overly large visual token set that significantly hinders inference efficiency under constrained token budgets.
Visual token pruning has emerged as a prevalent strategy to address this issue.
Nevertheless, most existing pruners are primarily tailored to 2D inputs or rely on indirect geometric cues, which limits their ability to explicitly retain semantically critical objects and maintain sufficient spatial coverage for robust 3D reasoning.
In this paper, we propose \name, a semantic-aware and geometry-guided token reduction framework for efficient 3D QA with multi-view images.
Specifically, \name first preserves semantically salient tokens through an attention-based importance module (\imp), ensuring that object-critical evidence is retained.
It then complements these tokens with spatially diverse ones via a geometry-guided selector (\diversity), which jointly considers semantic relevance and 3D geometric distance.
This cooperation between saliency preservation and geometry-guided diversification balances object-level evidence and global scene coverage under aggressive token reduction.
Extensive experiments on ScanQA and OpenEQA demonstrate that \name substantially improves inference efficiency, reducing the visual token budget by 91\% and inference latency by 86\%, while maintaining competitive performance in 3D reasoning tasks.
\end{abstract}
\begin{document}
\maketitle

\begin{figure}[t]
    \centering
    \begin{overpic}[width=1.0\linewidth]{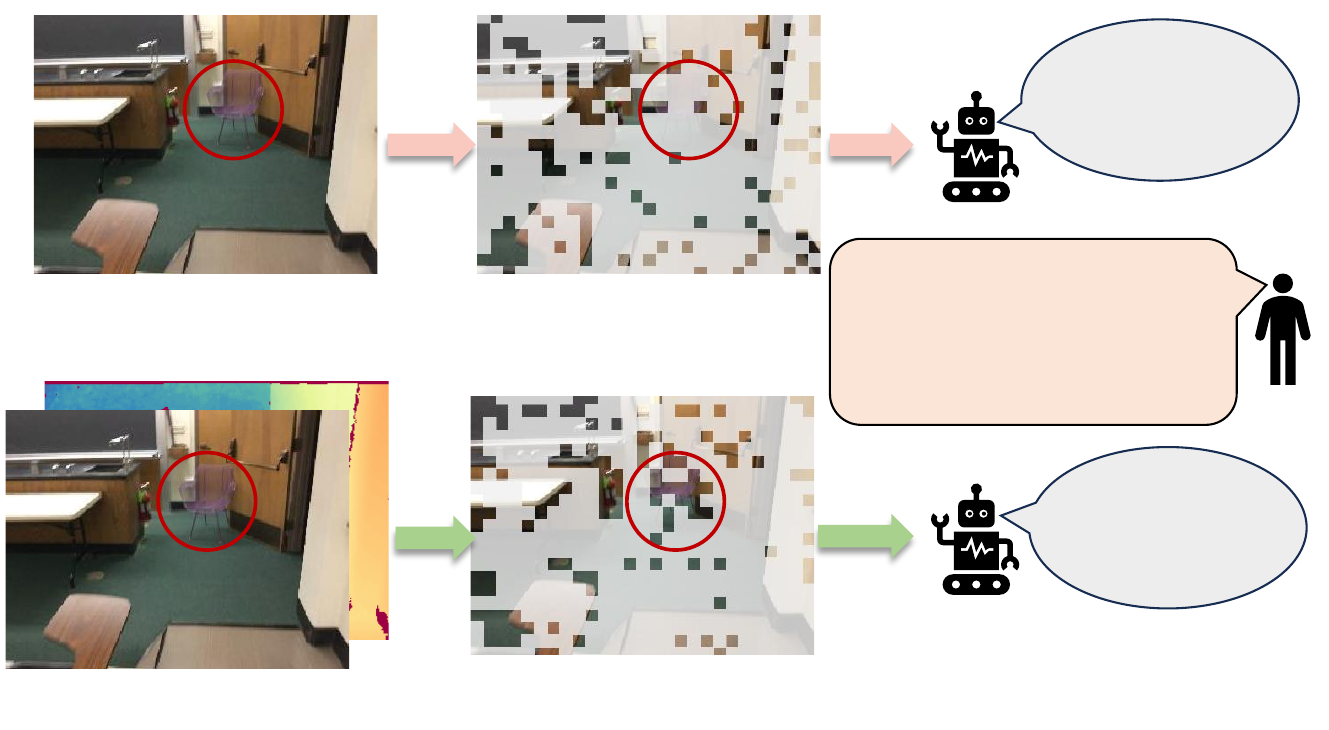}
        \small
        \put(16, 30){\makebox[0pt]{Image}}       
        \put(50, 30){\makebox[0pt]{VisPruner~\cite{vispruner}}}
        \put(79, 46){A: \textit{black}} 
        \put(92, 45){\textcolor{red}{\Large\ding{55}}}
        \put(63, 29){\parbox{2.6cm}{
            Q: \scriptsize{\textit{What color chair is to the right of the white table?}}
        }}
        \put(15, 0){\makebox[0pt]{RGB-D}} 
        \put(50, 0){\makebox[0pt]{Ours}}
        \put(79, 14){A: \textit{\color{purple}{purple}}}
        \put(94, 13){\textcolor{green!60!black}{\Large\ding{51}}} 
    \end{overpic}
    % \vspace{0.5em} 
    \caption{\textbf{Comparison between 2D visual token pruning and spatially-aware pruning for 3D QA.} 
    The visualization overlays retained visual tokens on the input image, where transparent regions indicate retained tokens and masked regions denote discarded ones.
    Conventional 2D pruning methods (top) allocate excessive tokens to background regions due to the lack of spatial awareness. In contrast, integrating spatial information (bottom) encourages more object-centric attention and more uniform sampling in 3D space, preserving richer and more comprehensive scene content for downstream reasoning.}
    \label{fig:fig1}
\end{figure}

\section{Introduction}
% \IEEEPARstart{U}{nderstanding}
% % Understanding 
% and reasoning about 3D environments is a fundamental capability for intelligent systems interacting with the physical world.
% 3D Question Answering (3D QA)~\cite{scanqa, openeqa, dspnet} is a multimodal task that aims to answer questions about 3D scenes.
The ability to understand and reason about 3D environments is crucial for intelligent systems interacting with the physical world.
3D Question Answering (3D QA)~\cite{scanqa, openeqa, dspnet} formulates this capability as a multimodal task of answering language queries grounded in 3D scenes,
it requires both semantic understanding to interpret scene context, and spatial reasoning to comprehend geometric relationships.
Consequently, 3D QA has been widely adopted across a range of real-world applications, including embodied intelligence~\cite{LI2026103624, huang2024embodied, savva2019habitat}, robotic navigation and interaction~\cite{yin2024sg, zhen20243d}, and autonomous driving~\cite{qian2024nuscenes, sun20243d}.

Applying vision-language models (VLMs) to 3D QA has become an emerging research direction~\cite{openeqa, 3d-llm, bridging, dtc, cdview} with the booming of VLMs~\cite{llava-onevision, llava-next, video-3d, qwen2,zhao2025open}. 
% Existing methods can be broadly categorized into three groups.
Early two-stage approaches~\cite{openeqa, enhancing} attempted to leverage VLMs to generate textual descriptions of 3D scenes and feed these descriptions into large language models for question answering. 
However, two-stage pipelines compress rich visual information into short text, which often discards fine-grained details. 
More recent 3D-input methods~\cite{3UR-LLM, SIG3D, gpt4point++, 3d-llm, splattalk} directly incorporate 3D data as visual inputs, enabling explicit utilization of geometric information. 
Despite this advantage, such methods often suffer from the scarcity and limited diversity of 3D training data, making it difficult to train robust 3D-VLMs.
Another line of work~\cite{bridging,dtc, cdview} replaces 3D data with multi-view images for 3D QA, building on the strong performance of multi-frame 2D VLMs across various tasks~\cite{Multi-spatialmllm, cheng20253d,Mono3DVLT, cogvlm2, llava-next, llavanext-video}.
In contrast to scarce 3D training data, multi-view images can be readily captured in practice~\cite{wu2023leveraging}. 
% Consequently, 2D pretrained VLMs trained on abundant data exhibit stronger generalization performance compared to direct 3D-input methods.
As a result, multi-view image-based 3D QA with pre-trained 2D VLMs has become increasingly popular. This trend motivates us to explore how to efficiently leverage multi-view visual information within pre-trained 2D VLMs for 3D QA.

Multi-view approaches have shown promising results by leveraging pre-trained 2D VLMs. 
However, aggregating multi-views inevitably introduces substantial visual redundancy, producing an overly long visual token sequence that significantly hinders inference efficiency under limited token budgets.
To alleviate these constraints, token reduction methods designed for 2D VLMs, such as ToMe~\cite{tome} and VisPruner~\cite{vispruner}, effectively reduce memory consumption. 
However, these methods are designed for 2D inputs and do not model 3D spatial structure.
When applied to 3D QA using VLMs, they often lack 3D spatial awareness, limiting their effectiveness in multi-view reasoning.
Recent studies have begun to incorporate 3D information into token reduction strategies.
Some methods adopt image retrieval strategies~\cite{bridging, cdview} to reduce the number of input images, while others employ token pruning~\cite{dtc} or token merging~\cite{tosa} techniques to reduce the number of visual tokens.
Nevertheless, image retrieval strategies~\cite{bridging, cdview} do not explicitly leverage 3D geometry, they reduce visual tokens by simply decreasing the number of input images, which can still leave substantial redundancy in the remaining tokens.
Moreover, while 3D-aware pruning~\cite{dtc} and merging~\cite{tosa} typically use 3D cues as auxiliary signals, they do not explicitly preserve salient visual tokens and ensure diverse spatial coverage of the scene during reduction. This may lead to the removal of critical information and insufficient coverage of key regions, potentially degrading the accuracy of answering. Consequently, it is crucial to design token reduction strategies that not only preserve semantically important tokens but also guarantee diverse spatial coverage across the scene.

% Overall, token reduction for 3D QA faces two key challenges: 
% (1) Existing 3D-aware approaches~\cite{dtc, tosa} may discard salient object information under aggressive reduction.
% (2) 
To address these challenges, we propose \name for 3D question answering. As illustrated in Fig.~\ref{fig:fig1}, 
\name leverages 3D geometric priors to reduce redundant multi-view tokens while preserving essential visual cues and spatial diversity.
To prevent the loss of information regarding primary objects during the reduction process, we introduce \imp. 
This module identifies and retains tokens corresponding to principal objects based on their importance estimated from attention scores.
After preserving salient tokens for semantically critical objects, we further introduce \diversity to enhance the scene representation by capturing rich contextual and fine-grained scene details.
Specifically, \diversity back-projects the remaining candidate tokens into a unified 3D coordinate space using camera extrinsic parameters and depth maps, and then selects spatially diverse tokens based on a joint semantic-spatial metric that combines feature similarity with 3D distance.
In summary, our contributions are three-fold:
% \begin{itemize}
%     \item We propose \name, a novel plug-and-play token reduction module that significantly reduces the number of visual tokens in existing 2D vision–language models for 3D QA.
%     \item We develop a cooperative token selection strategy that balances saliency preservation and spatial coverage. Specifically, we first retain subject-critical tokens via attention-based ranking, and then select spatially diverse tokens using a geometry-aware module guided by semantic similarity and 3D proximity.
% \end{itemize}

\begin{itemize}
    \item We propose \name, a semantic-aware and geometry-guided token reduction framework that preserves salient tokens and supplements them with spatially diverse ones to reduce multi-view redundancy.
    \item We develop \imp, an attention-based importance module that retains tokens of semantically critical objects for object-centric 3D reasoning.
    \item We design \diversity, a geometry-guided selector that combines semantic similarity with 3D distance to ensure broad spatial coverage under aggressive reduction.
\end{itemize}

Experimental results demonstrate that \name achieves state-of-the-art performance on both the ScanQA~\cite{scanqa} and OpenEQA~\cite{openeqa} benchmarks. In particular, on ScanQA, \name retains only 23\% of the original visual tokens while achieving better performance than the full-token base model.

\section{Related Work}

\subsection{3D QA with Explicit 3D Representations}
Early 3D question answering (3D QA) approaches relied on explicit 3D scene representations to provide geometric priors for reasoning.
ScanQA~\cite{scanqa} pioneered the use of point clouds for 3D QA, and subsequent works such as DSPNet~\cite{dspnet} further combined point clouds with multi-view to enhance scene understanding.
With the emergence of large vision-language models (VLMs), several studies explored incorporating 3D representations into VLMs to promote 3D world understanding.
Among various 3D modalities, point clouds remain the most common representation~\cite{LEO, SIG3D, 10313987, 10354431, gpt4point++, Language-Assisted}.
Other works adopt reconstructed 3D scene representations, including implicit neural fields~\cite{3d-llm} and 3D Gaussian Splatting~\cite{splattalk}, to serve as visual inputs for language models.
While these methods effectively leverage explicit geometric information, their progress is often constrained by the limited scale and diversity of available 3D datasets.

\subsection{3D QA with 2D VLMs}
Recent studies have shown that multi-view images captured from 3D scenes can be directly fed into existing 2D VLMs to perform 3D question answering.
This paradigm benefits from large-scale 2D pretraining and avoids the need for explicit 3D representations.
Early approaches often followed two-stage pipelines, where image captions or scene descriptions were first generated and then fed into large language models (LLMs) for reasoning~\cite{openeqa, enhancing}, at the cost of significant visual information loss.
More recent multi-frame and video-based VLMs~\cite{llava-onevision, llava-next, video-3d} adopt modular architectures that decouple visual perception from language reasoning.
These models employ pre-trained 2D vision encoders to extract frame-level representations, while LLMs aggregate information across views to perform semantic and spatial reasoning.
Representative works~\cite{Video-chatgpt, Video-llama, videochat, moviechat} demonstrate that such architectures enable effective multi-view reasoning without explicit 3D inputs.
Despite their strong perceptual priors, VLM-based 3D QA methods face practical constraints related to context length and computational efficiency, motivating the need for effective visual token reduction.

\subsection{Token Reduction for VLMs}
Visual tokens in VLMs often exhibit substantial redundancy~\cite{10855481, attention, tfrnet}, particularly when processing multi-view inputs or long visual sequences.
Unlike textual tokens, which are highly compact and abstract, visual tokens preserve dense perceptual and spatial information, resulting in a large number of tokens with low effective information density. In multi-view settings, background regions, flat surfaces, repetitive textures, and visually similar objects are frequently over-represented across different viewpoints, leading to significant redundancy in the visual token space.

To address this issue, a growing body of work has explored training-free token reduction strategies for VLMs, which can be applied at inference time without modifying model parameters.
These methods are attractive due to their plug-and-play nature and can be broadly categorized into token pruning and token merging approaches.
% These methods mainly fall into two categories: token pruning and token merging.
Token pruning methods discard less informative tokens based on attention or importance estimation~\cite{fastv, vispruner}. A common strategy is to leverage attention signals as a proxy for token relevance, under the assumption that tokens receiving higher attention are more critical for downstream reasoning. Representative works such as VisPruner~\cite{vispruner} utilize cross-modal or self-attention distributions to identify and remove low-importance visual tokens.
On the other hand, token merging methods~\cite{tome, DToM} reduce the token count by fusing similar tokens during inference. Approaches such as ToMe~\cite{tome} merge tokens based on feature similarity, effectively compressing redundant visual representations while preserving global structural information.
% Although effective in reducing computation and memory costs, most existing token reduction methods operate purely in the 2D domain. Consequently, they cannot exploit 3D spatial information for 3D QA tasks, leading to suboptimal reasoning performance.
Although effective in reducing computation and memory costs, most existing token reduction methods operate purely in the 2D domain. Consequently, these methods cannot exploit 3D spatial information when applied to 3D QA tasks, where multi-view redundancy across viewpoints play a critical role, leading to suboptimal reasoning performance.

\begin{figure*}[!htb]
    \centering
    \includegraphics[width=0.98\textwidth]{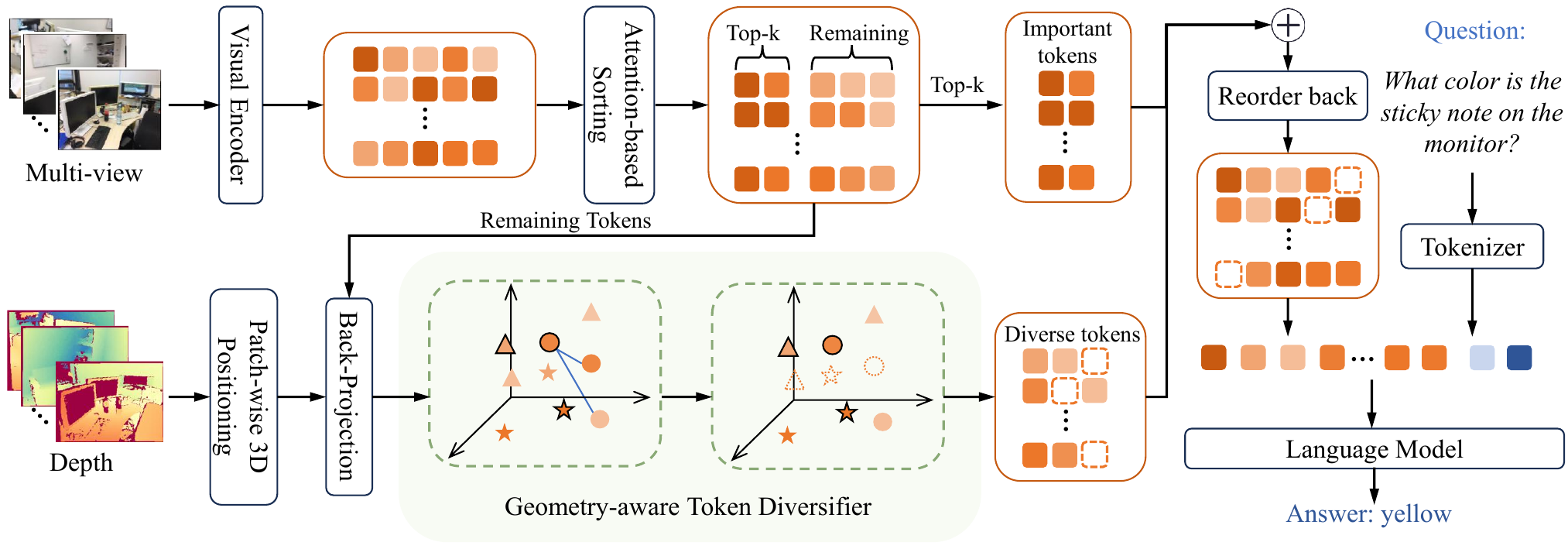}
    \caption{\textbf{Overall framework of our method.} Our module is inserted between the visual encoder and the LLM and consists of three components: (1) \textbf{3D-Aware Feature Construction}, (2) \textbf{Salient Token Selection}, and (3) \textbf{Diverse Token Selection}. Initially, the visual encoder extracts 2D visual tokens from multi-view inputs, where darker colors indicate higher attention scores. We select the top-$k$ tokens as \textit{important tokens}. To preserve spatial context, the remaining tokens are back-projected into 3D space using the corresponding depth map and processed by the \textbf{\diversity} to obtain \textit{diverse tokens}. In the \diversity, tokens sharing the same shape originate from the same input view. Dashed outlines denote tokens discarded during selection. Finally, the important and diverse tokens are concatenated and fed into the LLM for cross-modal reasoning.}
    \label{fig:placeholder}
\label{fig:pipline}
\end{figure*}

To bridge this gap, recent works have begun incorporating geometric cues into token reduction.
Image retrieval-based approaches~\cite{bridging, cdview} leverage camera parameters to select a subset of informative views for reducing redundancy, but may still retain significant token-level overlap.
Moreover, these approaches require training task-specific view selection modules, which limits their applicability to off-the-shelf pre-trained VLMs.
DTC~\cite{dtc} and ToSA~\cite{tosa} further integrate depth and camera information to perform spatially informed token reduction. 
However, these approaches either treat 3D cues as auxiliary signals or rely primarily on semantic similarity, without explicitly balancing object-level saliency preservation and spatial diversity across views.
As a result, they may lose semantically important information while providing insufficient coverage of spatially distributed objects.
Therefore, an open challenge remains to design a token reduction strategy that jointly preserves salient visual semantics and broad spatial coverage for robust 3D question answering.

% \section{Method}
\section{Methodology}
\label{sec:method}
Our design is motivated by the observation that effective token reduction for 3D question answering must satisfy two complementary objectives:
(1) preserving tokens that correspond to semantically critical objects, and
(2) maintaining broad spatial coverage to support global scene understanding.
Accordingly, we perform token selection in a 3D-aware manner and decompose it into two cooperative stages: salient token preservation and geometry-guided diverse token selection.
The overall pipeline of our method is illustrated in Fig.~\ref{fig:pipline}. 
\subsection{3D-Aware Feature Construction}
\label{sub:spatial}
To capture the geometric structure of a 3D scene, we project 2D visual features into a unified 3D coordinate space using multi-view images and their associated depth maps. 
This process yields a 3D-aware scene representation for subsequent token selection. 
Inspired by DTC~\cite{dtc}, we assign patch-level 3D coordinates to visual tokens as a lightweight geometric abstraction. 
This design provides sufficient spatial cues while avoiding the overhead of dense point-level representations.
Specifically, for each input image, we first feed it into the visual encoder to extract a visual feature map $f \in \mathbb{R}^{N \times D}$,  where \(N\) denotes the number of image patches and \(D\) denotes the feature dimension. Given the depth map and camera pose of the image, each patch is back-projected into 3D space. Let $d(p)$ denote the per-pixel depth, $K$ the intrinsic matrix, and $T = [R\; t] \in SE(3)$ the camera extrinsic transformation from the camera to the world coordinate system. For the \(i\)-th patch with pixel set \(\Omega_i\), its 3D location is computed as the average of the world coordinates of all pixels within that region:
\begin{equation}
\label{eq:coordinates}
    \mathbf{c}_i = \frac{1}{|\Omega_i|} 
    \sum_{p \in \Omega_i} \Pi(d(p), K^{-1}, T),
\end{equation}
where \(\Pi(\cdot)\) denotes the back-projection operation, and \(|\Omega_i|\) is the pixel count of the \(i\)-th patch region. For each view \( v \), we obtain an explicit 3D coordinate \(\mathbf{c}_i^{(v)}\) for every token \(f_i^{(v)}\). To construct a globally consistent 3D representation, all tokens from different views are represented in a unified world coordinate frame using their depth maps and camera poses:
\begin{equation}
\label{eq:scene}
    % \mathcal{F}_{3D} = \{(f_i, \mathbf{c}_i)\}_{i=1}^{N}
    \mathcal{F} = \bigcup_{v=1}^{V} \{ (f_i^{(v)}, \mathbf{c}_i^{(v)}) \}_{i=1}^{N}.
\end{equation}
This allows tokens from different views to be represented in a unified world coordinate frame, enabling cross-view spatial comparison.

\subsection{Selecting Salient Tokens}
\label{sub:imp}
Attention scores have been widely adopted as an effective measure of token importance in various vision and multimodal tasks~\cite{vispruner, fastv, xing2024pyramiddrop, adav}.  
Intuitively, tokens receiving higher aggregated attention tend to correspond to visually salient objects or regions that are more relevant to downstream reasoning, making attention a practical proxy for token importance.

Following VisPruner~\cite{vispruner}, which directly uses the attention distribution from the visual encoder’s \texttt{[CLS]} token to all other visual tokens as an importance measure, we adapt this strategy to vision encoders that do not contain a \texttt{[CLS]} token. In this case, we compute token importance by averaging the attention values along the column dimension of the attention matrix $A \in \mathbb{R}^{N \times N}$. Here, \(A\) denotes the self-attention matrix from the last block of the visual encoder. Specifically, the importance of the \(j\)-th token is determined by the average amount of attention it receives from all other tokens:
\begin{equation}
    s_j = \frac{1}{N} \sum_{i=1}^{N} A_{ij},
\label{eq:attention}
\end{equation}
where \(N\) denotes the number of image patches and \( A_{ij}\) denotes the attention weight from token \(i\) to token \(j\). 
Based on the attention scores \(\mathcal{S} = \{s_j\}_{j=1}^N\), we sort all visual tokens in descending order of their scores and obtain an ordered index sequence $(k_1, \dots, k_N)$, where $s_{k_1} \ge s_{k_2} \ge \dots \ge s_{k_N}$. 
Given a predefined important-token ratio \(r\) and a target retention budget of \(M\) tokens, we then select the top \( n=\lfloor rM \rfloor \) indices as \(\mathcal{I} = \{k_1, \dots, k_n\}\), where \(\lfloor \cdot \rfloor\) denotes the floor operation.

\subsection{Selecting Diverse Tokens}
\label{sub:div}
\begin{figure}[t]
    \centering
    \begin{overpic}[width=1.0\linewidth]{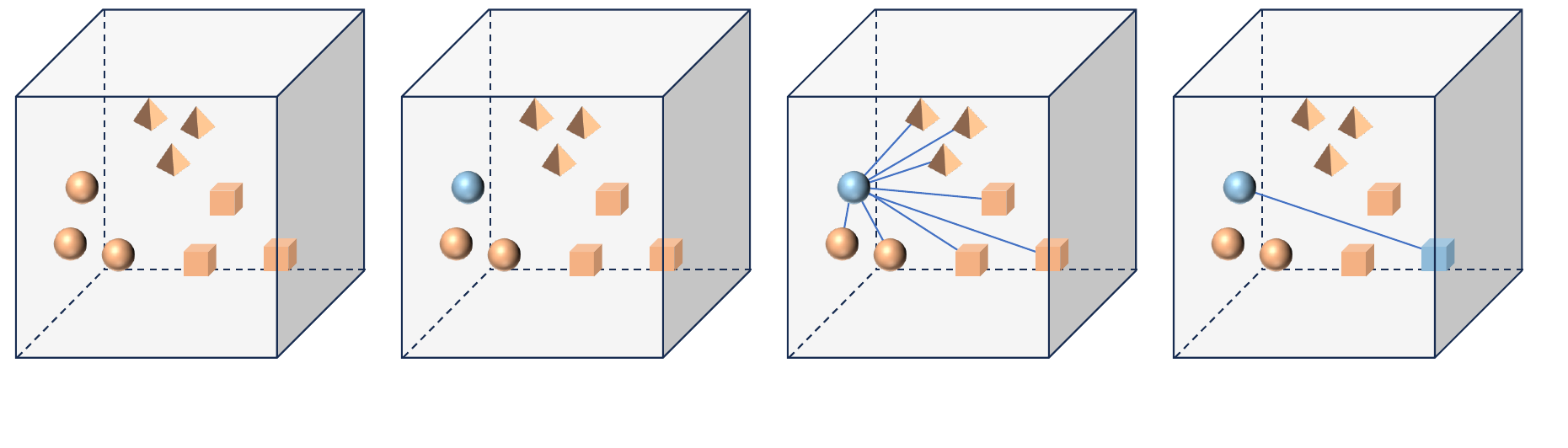}
        \small
        \put(28, 18){\parbox[c]{0.2\linewidth}{\footnotesize
            \(r_0\)
        }}
        \put(52, 18){\parbox[c]{0.2\linewidth}{\footnotesize
            \(r_0\)
        }}
        \put(77, 18){\parbox[c]{0.2\linewidth}{\footnotesize
            \(r_0\)
        }}\
        \put(93, 13){\parbox[t]{0.2\linewidth}{\footnotesize
            \(r_1\)
        }}

        \put(3, 0){\parbox[t]{0.25\linewidth}{\footnotesize
            Initialization
        }}
        \put(28, 0){\parbox[t]{0.2\linewidth}{\footnotesize
            Select \(r_0\)
        }}
        \put(53, 0){\parbox[t]{0.25\linewidth}{\footnotesize
            Compute \(d^{geo}\)
        }}
        \put(80, 0){\parbox[t]{0.2\linewidth}{\footnotesize
            Select \(r_1\)
        }}
    \end{overpic}
    % \vspace{0.5em} 
    \caption{\textbf{Illustration of \diversity in 3D space.}
    The outer cube represents the 3D scene space, and geometric primitives denote visual tokens extracted from a single image. We first initialize the selected set \(\mathcal{D}\) with the highest-attention token \(r_0\). Subsequently, we iteratively calculate the minimum fusion distance from the remaining tokens to \(\mathcal{D}\) and select the farthest candidate \(r_1\) to update the set. This strategy effectively reduces redundancy by discouraging tokens that are spatially or semantically close to those already selected.
    }
    \label{fig:fps}
\end{figure}
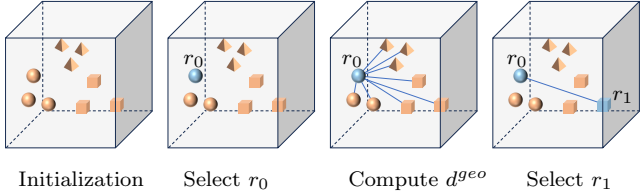

To prevent the model from over-focusing on a few local regions under high reduction ratios and improve scene coverage and global understanding, we sample spatially diverse tokens from the remaining features after selecting the salient tokens.
We are inspired by Farthest Point Sampling (FPS)~\cite{pointnet++}, which was originally designed for point clouds and encourages approximately uniform coverage.
Different from dense point clouds, our candidates are visual tokens from discrete image patches. 
We define a unified metric that combines normalized Euclidean distance with semantic similarity between tokens. This metric guides diverse token selection in a semantic–spatial fusion space.
It encourages broad spatial coverage while avoiding redundant tokens with high visual similarity. 
The overall \diversity process is illustrated in Fig.~\ref{fig:fps}. 

% \begin{figure}[!htb]
% \centering
% \includegraphics[width=\columnwidth]{figs/fps.pdf}
% \caption{\textbf{Illustration of attention-guided semantic–spatial token selection in 3D space.}
% The outer cube represents the 3D scene space, and geometric primitives denote visual tokens extracted from a single image. We first initialize the selected set \(\mathcal{D}\) with the highest-attention token \(r_0\). Subsequently, we iteratively calculate the minimum fusion distance from the remaining tokens to \(\mathcal{D}\) and select the farthest candidate \(r_1\) to update the set. This strategy effectively reduces redundancy by discouraging tokens that are spatially or semantically close to those already selected.
% }
% \label{fig:fig3}
% \vspace{-4mm}
% \end{figure}

We initialize the diverse set with the highest-attention token among the remaining candidates.
Specifically, given the remaining index sequence \(\mathcal{R} = (k_{n+1}, \dots, k_N)\)sorted in descending order of attention scores, we set \(\mathcal{D} = \{k_{n+1}\}\). We then iteratively expand \(\mathcal{D}\) to the target size by repeating the following steps:
\subsubsection{Fusion Distance Computation}
    For each candidate token $f_r$ with $r \in \mathcal{R} $, we compute a semantic–spatial distance to the current diverse set \(\mathcal{D}\). For any token pair \((r, j)\) with \(j \in \mathcal{D}\), we define:\\
    \noindent{Geometric distance:}
    \begin{equation}
    \label{eq:geo}
    \quad d_{rj} = \lVert \mathbf{c}_r - \mathbf{c}_j \rVert_2;
    \end{equation}
    
    \noindent{Semantic similarity:}
    \begin{equation}
    \label{eq:smi}
    % \quad s_{rj} = \mathbf{f}_r^\top \mathbf{f}_j;
    \quad s_{rj} = \frac{f_r^\top f_j}{\lVert f_r\rVert_2 \cdot \lVert f_j\rVert_2};
    \end{equation}
    
    \noindent{Semantic-spatial distance:}
    \begin{equation}
    \label{eq:ss}
    d^{\mathrm{geo}}_{rj}
    = \lambda \cdot \frac{{d_{rj}}}{d_x} + (1-\lambda)(1 - s_{rj}),
    \end{equation}
    where $d_x = \max_{r \in \mathcal{R}} \lVert \mathbf{c}_r - \mathbf{c}_j \rVert_2$
    is the maximum geometric distance in the first iteration, \(\lVert \cdot\rVert_2\) denotes the \(\ell_2\)-norm,
    and $\lambda \in [0,1]$ balances spatial and semantic terms. This formulation allows us to penalize both spatial proximity and semantic redundancy, encouraging tokens that are not only far apart in 3D space but also complementary in visual content.
    
    We define the distance from $f_r$ to the set \(\mathcal{D}\) as:
    \begin{equation}
    \label{eq:dr}
        d_r = \min_{j\in \mathcal{D}} d^{\mathrm{geo}}_{rj}.
    \end{equation}
\subsubsection{Farthest-point update}
    The next sampled index is chosen as:
    \begin{equation}
    \label{eq:r*}
        r^{*} = \arg\max_{r \in \mathcal{R}} d_r ,
    \end{equation}
    and the diverse set is updated:
    \begin{equation}
    \label{eq:update}
        \mathcal{D} \leftarrow \mathcal{D} \cup \{ r^{*} \}.
    \end{equation}
\subsubsection{Iterative sampling}
    Repeat Steps 1-2 until \(\lvert \mathcal{D} \rvert = M - \lfloor rM \rfloor\).
    
% The overall attention-guided semantic-spatial sampling process is illustrated in Fig.~\ref{fig:fps}. 
By combining attention-guided initialization with spatial-distribution and semantic-similarity constraints, this sampling strategy ensures that the selected diverse tokens maintain broad spatial coverage in 3D space. Consequently, these tokens complement the important tokens by capturing scene regions that would otherwise be overlooked, thereby providing richer and more comprehensive visual cues for downstream reasoning. The pipeline of \diversity can be seen in Algorithm ~\ref{alg:diverse}.

\subsection{Inference Procedure}
During inference, we discard the remaining visual tokens and use only the union of the important and diverse tokens to replace the original visual token sequence.
The final ordered token sequence is defined as:
    \begin{equation}
        \mathcal{F}_{\text{final}} = \{ f_k \mid k \in \operatorname{sort}(\mathcal{I} \cup \mathcal{D}) \},
        % \mathcal{F}_{\text{final}} = [\, f_k \,]_{k \in \operatorname{sort}(\mathcal{I} \cup \mathcal{D})},
    \end{equation}
% where $\mathrm{sort}(\cdot)$ restores the original token order according to the token indices in the visual encoder.
where $\operatorname{sort}(\cdot)$ restores the original ordering of tokens based on their indices in the visual encoder. This preserves the original token order expected by the LLM.
The reduced token set $\mathcal{F}_{\text{final}}$ is then fed into the language model (LLM) together with the input query $q$ for cross-modal reasoning:
    \begin{equation}
    \label{eq:llm}
        \hat{y} = \mathrm{LLM}(\mathcal{F}_{\text{final}},\, q),
    \end{equation}
where $\hat{y}$ denotes the generated textual answer produced by the LLM.
By preserving both semantic importance and spatial diversity, the proposed token selection strategy significantly reduces the number of visual tokens required during inference, while maintaining strong scene understanding and reasoning capability.

\begin{algorithm}[t]
\renewcommand{\algorithmicrequire}{\textbf{Input:}}
\renewcommand{\algorithmicensure}{\textbf{Output:}}
\caption{\diversity}
\label{alg:diverse}
\begin{algorithmic}[1]
\REQUIRE Remaining visual token index sequence \(\mathcal{R}\) with 3D coordinates and features, target number of diverse tokens \(n_{div}\)
\ENSURE Diverse token index set $\mathcal{D}$
\STATE \textbf{Step 1: Initialization }
\STATE Initialize the diverse set \(\mathcal{D}\) with the token \(r_0\) that has the highest attention score in \(\mathcal{R}\); 

\STATE \textbf{Step 2: Iterative Diverse Token Selection}
\WHILE{$|\mathcal{D}| < n_{div}$}
    \STATE For each candidate token \(r \in \mathcal{R}\), compute its minimum semantic-spatial distance to the current diverse set \(\mathcal{D}\) according to Eq.~(\ref{eq:geo})--(\ref{eq:dr});

\STATE Select the farthest candidate \(r^*\) according to Eq.~(\ref{eq:r*}); 

\STATE Update the diverse set \(\mathcal{D}\) with \(r^*\)

\ENDWHILE

\RETURN $\mathcal{D}$
\end{algorithmic}
\end{algorithm}

\section{Experiments}
\label{sec:exp}

% \subsection{Experimental Setup}
\subsection{Implementation Details}
% \noindent{\textbf{Implementation Details.}}
We adopt LLaVA-OneVision-7B (OV)~\cite{onevison} as the VLM for all experiments, following prior work~\cite{dtc, cdview, tosa}, and keep the model frozen throughout. To capture the spatial information of each 3D scene, we uniformly sample 12 RGB images from different viewpoints as visual inputs, following a multi-view sampling strategy similar to DTC~\cite{dtc}. All images are resized to \(384 \times 384\) and fed into SigLIP~\cite{siglip} to extract multi-view visual tokens, resulting in 8,748 visual tokens per scene before token selection.
Our proposed \name performs token reduction allowing the number of retained tokens to be flexibly adjusted according to the inference budget. It is applied after visual encoding and before the tokens are fed into the LLM, without modifying the LLM architecture. The depth maps used by \name are obtained from dataset annotations and are only employed to project visual tokens into 3D space during the token selection process.
In all experiments, the balancing parameter \(\lambda\) is fixed at 0.5 to trade off spatial distance and semantic relevance. We observe that under aggressive token reduction, emphasizing token diversity promotes broader spatial coverage and reduces the risk of over-concentrating on a few salient regions. Accordingly, the importance ratio is adjusted according to the retention ratio to balance important and diverse tokens.

% \noindent{\textbf{Datasets.}}
To comprehensively evaluate the effectiveness of our method for 3D question answering (3D QA), we conduct experiments on two representative benchmarks, ScanQA~\cite{scanqa} and OpenEQA~\cite{openeqa}, which are widely used in prior work~\cite{dtc, dspnet, 3d-llm}. 
ScanQA is built upon the large-scale indoor scene dataset ScanNet~\cite{scannet}, and contains approximately 8,000 real-world indoor scenes and over 41,000 question–answer pairs. It primarily evaluates a model’s understanding of scene geometry and semantics, including object localization, relational reasoning, and cross-view consistency. In contrast, OpenEQA targets the more challenging embodied question answering (EQA) setting and comprises more than 1,600 high-quality, human-annotated question–answer pairs collected from the ScanNet~\cite{scannet} and HM3D~\cite{hm3d} datasets. The questions span more than 180 real-world scenes and cover multiple categories, such as object recognition, attribute understanding, spatial reasoning, and interaction comprehension, thereby providing a rigorous evaluation of a model’s multimodal reasoning and generalization abilities in complex 3D environments.

\begin{table}[!t]
    \centering
    \caption{Comparison of our base model with representative 3DQA methods on ScanQA.}
    % \resizebox{0.85\linewidth}{!}{
    \begin{tabular}{l @{\hspace{1.5cm}} c}
    \toprule Model & EM@1\\
    \midrule
    \textit{Specialized Models} & \\
    VoteNet+MCAN~\cite{scanrefer}& 17.3\\
    ScanQA~\cite{scanqa} & 21.1 \\
    3D-VisTA~\cite{3d-vista} & 22.4 \\
    DSPNet~\cite{dspnet} & 23.5 \\
    BridgeQA~\cite{bridging} & 27.0 \\
    
    \midrule
    \textit{Video-LMMs} & \\
    LLaVA-NeXT-Video~\cite{llavanext-video} & 18.7\\
    VideoChat2~\cite{videochat}& 19.2\\
    MovieChat (w/ LLaVA-OV-7B)~\cite{moviechat} & 26.0\\
    
    \midrule
    \textit{Task-specialized 3D-LMMs} & \\
    3D-LLM~\cite{3d-llm} & 20.5 \\
    FE-3DGQA~\cite{FE-3DGQA} &22.3\\
    LEO~\cite{LEO}  & 24.5 \\
    UniVLG~\cite{jain2025unifying} & 25.7 \\
    Scene-LLM~\cite{Scene-llm} & 27.2 \\
    \midrule
    \textit{Our Base Model} & \\
     \textbf{LLaVA-OV-7B}~\cite{onevison} & \textbf{27.6} \\

    \bottomrule
    \end{tabular}
    \label{tab:models_scan}
    % }
\end{table}

\subsection{Baseline}
% \textit{Overall model comparison:}
\subsubsection{Overall model comparison}
We evaluate representative methods on both ScanQA and OpenEQA. This comparison serves to assess the competitiveness of our chosen base model.

We compare three categories of representative methods on the ScanQA dataset, as summarized in Table~\ref{tab:models_scan}.
The first category includes task-specific 3D-QA models~\cite{scanrefer, scanqa, 3d-vista, dspnet} that leverage explicit 3D information, such as point clouds reconstructed from RGB-D, to exploit geometric priors for structured scene understanding.
The video-based VLMs~\cite{llavanext-video, videochat, moviechat}, which process multi-frame inputs and implicitly capture 3D cues via temporal modeling.
The third category comprises task-specialized 3D-LMMs~\cite{3d-llm, FE-3DGQA, LEO, jain2025unifying, Scene-llm}, which adapt large vision-language models to 3D question answering by incorporating explicit 3D scene representations from multi-view, object-centric 3D tokens, or point-cloud-derived features, and fine-tuning/instruction-tuning them on 3D QA data.

For OpenEQA (Table~\ref{tab:models_open}), we compare several representative closed-source and open-source VLMs~\cite{Video-llama, openeqa, auroracap, Video-chatgpt, moviechat}.
All compared methods take multi-frame scene images as input, while some approaches~\cite{openeqa} additionally incorporate auxiliary information (e.g., scene graphs, captions, or sparse voxel maps) to enhance LLM-based reasoning.
Across both benchmarks, our base model (LLaVA-OneVision)~\cite{onevison} achieves competitive performance despite using neither explicit 3D inputs nor additional auxiliary information.
We thus use LLaVA-OneVision as the fixed base VLM and compare token reduction methods under identical settings for a fair evaluation.

% \textit{Comparison with token reduction methods:}
\subsubsection{Comparison with token reduction methods}
We further compare our approach with representative token reduction methods under different token retention ratios. Specifically, we consider DTC~\cite{dtc} as a 3D-aware token reduction method and VisPruner~\cite{vispruner} as a representative 2D token pruning approach. For fair comparison, all pruning methods are applied to the same base model and evaluated under identical settings.

\begin{table}[htbp]
    \centering
    % \caption{Performance comparisons with the state-of-the-art methods on the OpenEQA in terms of LLM-Match.}
    \caption{Comparison of our base model with representative 3DQA methods on OpenEQA}
    % \resizebox{0.98\linewidth}{!}{
    \begin{tabular}{l  c}
    \toprule Model & LLM-Match\\
    \midrule
    \textit{Proprietary Models} & \\
    Gemini 1.0 Pro Vision & 44.9 \\
    Claude-3.5 Sonnet & 48.7 \\
    GPT4-V (15 frames) & 54.6 \\
    GPT4-V (50 frames) & 55.3 \\
    \midrule
    \textit{Open Models} & \\
    Video-LLaMA~\cite{Video-llama} & 20.0 \\
    LLaMA-2 w/ Concept Graph~\cite{openeqa} & 28.7 \\
    AuroraCap~\cite{auroracap} & 28.9 \\
    Video-ChatGPT~\cite{Video-chatgpt} & 32.1 \\
    LLaMA-2 w/ Sparse Voxel Map~\cite{openeqa} & 34.3 \\
    LLaMA-2 w/ LLaVA-1.5 caption~\cite{openeqa} & 36.8 \\
    MovieChat (w/ LLaVA-OV-7B)~\cite{moviechat} & 54.9 \\
    \midrule
    \textit{Our Base Model} & \\
     \textbf{LLaVA-OV-7B}~\cite{onevison} & \textbf{56.2} \\
    \bottomrule
    \end{tabular}
    \label{tab:models_open}
    % }
\end{table}

% \noindent{\textbf{Evaluation Metrics.}} 
\subsection{Evaluation Metrics} 
Following prior work~\cite{scanqa, dtc, cdview, dspnet}, we use Exact Match (EM@1) as the primary metric on ScanQA~\cite{scanqa}. EM@1 measures the percentage of predictions that exactly match the ground-truth answers, providing an objective and reproducible evaluation for open-ended question answering. 
For OpenEQA~\cite{openeqa}, we adopt LLM-Match, the metric introduced in the original OpenEQA paper. LLM-Match leverages a LLM to assess the semantic consistency between predicted and reference answers, which better reflects semantic correctness in embodied question answering scenarios. We follow the official OpenEQA evaluation protocol and adopt the provided LLM evaluation prompt with the GPT-4 checkpoint (gpt-4-1106-preview).

\begin{table}[!t]
    \centering
    \caption{Performance comparison of different pruning methods on ScanQA.}
    \resizebox{0.98\linewidth}{!}{
    \begin{tabular}{l c c c}
        \toprule Token Retention Ratio & DTC~\cite{dtc} & VisPruner~\cite{vispruner} & Ours \\
        \midrule
        \barpercent{100} & 27.6 & 27.6 &27.6 \\
        \barpercent{54} & 27.8{\scriptsize\textcolor{green!60!black}{(+0.2)}}& 
        28.3{\scriptsize\textcolor{green!60!black}{(+0.7)}}&
        \textbf{28.7}{\scriptsize\textcolor{green!60!black}{(+1.1)}}\\
        
        \barpercent{40} & 27.7{\scriptsize\textcolor{green!60!black}{(+0.1)}}& 
        27.7{\scriptsize\textcolor{green!60!black}{(+0.1)}}&
        \textbf{28.5}{\scriptsize\textcolor{green!60!black}{(+0.9)}}\\
        
        \barpercent{23} & 27.7{\scriptsize\textcolor{green!60!black}{(+0.1)}}& 
        26.6{\scriptsize\textcolor{red!90!black}{(-1)}}&
        \textbf{28.0}{\scriptsize\textcolor{green!60!black}{(+0.4)}}\\
        
        \barpercent{14} & \textbf{26.7}{\scriptsize\textcolor{red!90!black}{(-0.9)}}& 
        25.3{\scriptsize\textcolor{red!90!black}{(-2.3)}}&
        \textbf{26.7}{\scriptsize\textcolor{red!90!black}{(-0.9)}} \\
        
        \barpercent{9} & 26.1{\scriptsize\textcolor{red!90!black}{(-1.5)}}&
        23.2{\scriptsize\textcolor{red!90!black}{(-4.4)}}&
        \textbf{26.3}{\scriptsize\textcolor{red!90!black}{(-1.3)}}\\
        
        \bottomrule
    \end{tabular}
    \label{scan}
    \label{tab:scanqa}
    }
\end{table}

\subsection{Experimental Results and Discussion}
To examine how token reduction affects spatial understanding, we evaluate performance on ScanQA~\cite{scanqa} under different visual-token retention ratios. The questions in ScanQA often require identifying and localizing specific objects, making the performance particularly sensitive to the loss of object-centric tokens.
Table~\ref{tab:scanqa} compares our method with DTC~\cite{dtc} and VisPruner~\cite{vispruner}. 
DTC incorporates depth information, while VisPruner operates purely in the 2D domain.
Compared to DTC, which relies mainly on geometric cues, our method jointly considers token saliency and spatial diversity, leading to more robust performance under aggressive reduction.
As shown in Table~\ref{tab:scanqa}, both our method and DTC improve model performance at moderate retention ratios (54\%, 40\%, and 23\%). 
This observation aligns with prior findings~\cite{hao2025principles, ryoo2021tokenlearner} that token reduction improves performance by removing redundant visual tokens from multi-view, resulting in a more compact and informative token set.
In contrast, VisPruner exhibits a performance drop when the retention ratio is reduced to 23\%, indicating its limited robustness under aggressive token reduction.
Our method consistently yields larger performance gains than DTC, indicating its superior ability to select semantically and spatially relevant regions.
Under extreme reduction, our approach remains particularly robust. Even at a retention ratio of 9\%, it preserves 95.3\% of the original model performance.
These results suggest that explicitly preserving spatially and semantically diverse tokens helps maintain sufficient scene coverage for reliable 3D reasoning under severe token reduction.

\begin{table}
    \centering
    \caption{Performance comparison of different pruning methods on OpenEQA.}
    \resizebox{0.98\linewidth}{!}{
    \begin{tabular}{l c c c}
        \toprule Token Retention Ratio & DTC~\cite{dtc} & VisPruner~\cite{vispruner} & Ours \\
        \midrule
        \barpercent{100} & 56.2 & 56.2 &56.2 \\
        \barpercent{56} & {55.3}{\scriptsize\textcolor{red!90!black}{(-0.9)}}& 
        56{\scriptsize\textcolor{red!90!black}{(-0.2)}}&
        \textbf{56.1}{\scriptsize\textcolor{red!90!black}{(-0.1)}}\\
        
        \barpercent{43} & 54.3{\scriptsize\textcolor{red!90!black}{(-1.9)}}& 
        \textbf{56.1}{\scriptsize\textcolor{red!90!black}{(-0.1)}}&
        55.9{\scriptsize\textcolor{red!90!black}{(-0.3)}}\\
        
        \barpercent{26} & 54.1{\scriptsize\textcolor{red!90!black}{(-2.1)}}& 
        55.0{\scriptsize\textcolor{red!90!black}{(-1.2)}}&
        \textbf{55.3}{\scriptsize\textcolor{red!90!black}{(-0.9)}}\\
        
        \barpercent{17} & 52.5{\scriptsize\textcolor{red!90!black}{(-3.7)}}& 
        \textbf{53.1}{\scriptsize\textcolor{red!90!black}{(-3.1)}}&
        \textbf{53.1}{\scriptsize\textcolor{red!90!black}{(-3.1)}} \\
        
        \barpercent{8} & 49.3{\scriptsize\textcolor{red!90!black}{(-6.9)}}&
        47.3{\scriptsize\textcolor{red!90!black}{(-8.9)}}&
        \textbf{49.9}{\scriptsize\textcolor{red!90!black}{(-6.3)}}\\
        
        \bottomrule
    \end{tabular}
    \label{tab:openeqa}
    }
\end{table}

We further evaluate on OpenEQA~\cite{openeqa}, an open-vocabulary embodied question answering benchmark built on real-world scanned environments. It evaluates whether a model can answer natural-language questions by integrating information from multi-step observations.
Table~\ref{tab:openeqa} summarizes the results on OpenEQA under varying token retention ratios.
At moderate retention ratios (56\% and 43\%), both our method and VisPruner achieve performance comparable to the base model, indicating an effective trade-off between efficiency and reasoning accuracy.
In contrast, VisPruner suffers a substantial performance drop at low retention ratios, suggesting that purely 2D token pruning is insufficient for embodied reasoning tasks that rely on spatial continuity.
Even under extreme reduction, our method remains robust. At an 8\% retention ratio, it outperforms VisPruner by 2.6 points and DTC by 0.6 points.
These results highlight the importance of preserving spatially diverse tokens for maintaining global scene understanding in embodied 3D question answering.

% \begin{figure}[!t]
% \centering
% \includegraphics[width=0.49\textwidth]{figs/ablation.pdf}
% \vspace{-7mm}
% \caption{\textbf{Qualitative ablation of token selection strategies.} (a) Full tokens without reduction.
% (b) Saliency-only selection (\imp) preserves salient objects but may miss fine-grained details like cables.
% (c) Diversity-only selection (\diversity) improves spatial coverage but can overlook semantically important regions on the tabletop.
% (d) Our method combines saliency and spatial diversity, retaining salient objects while preserving fine-grained structures and producing more continuous and complete structures for large objects such as the table.
% }
% \label{fig:ablation}
% \vspace{-4mm}
% \end{figure}
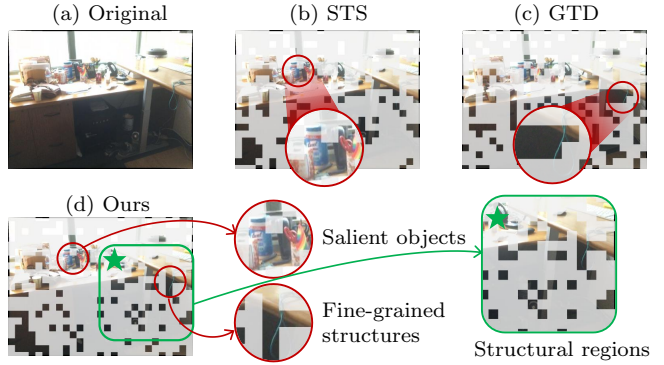
\begin{figure}[t]
    \centering
    \begin{overpic}[width=1.0\linewidth]{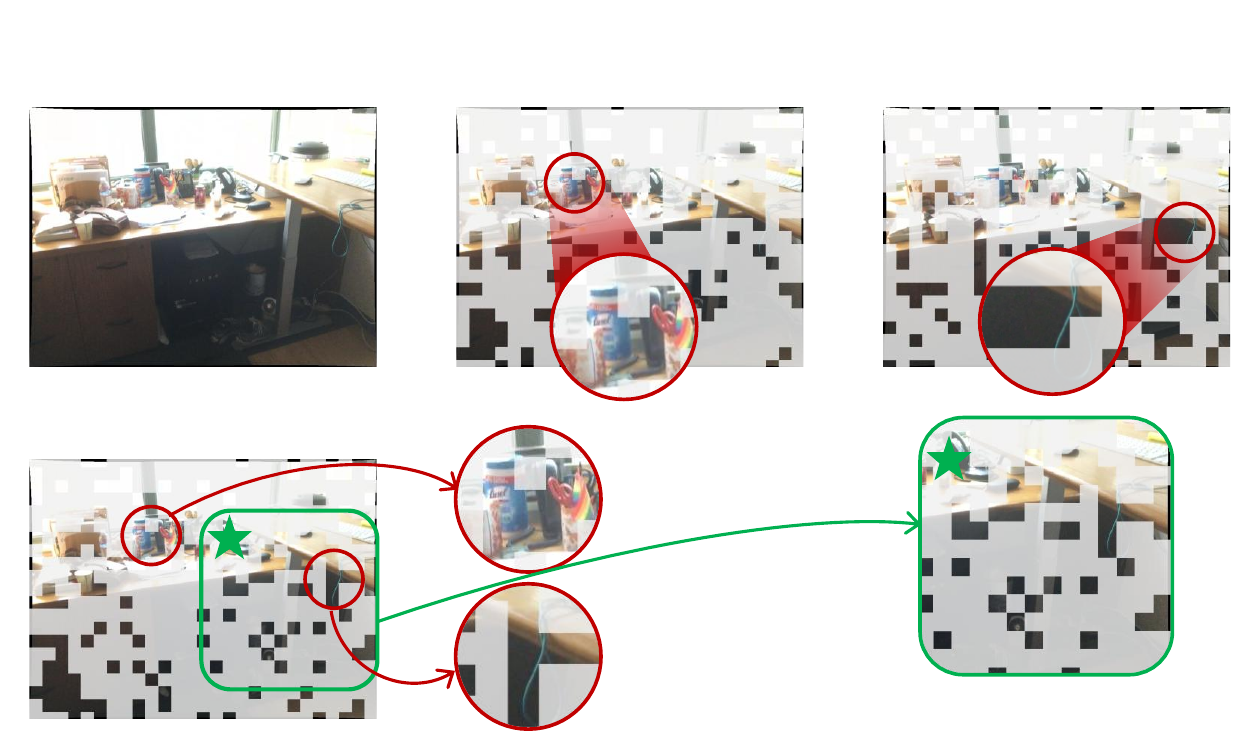}
    
        \put(9, 52){\parbox[t]{0.2\linewidth}{\footnotesize
            (a) Original
        }}
        \put(45, 52){\parbox[t]{0.2\linewidth}{\footnotesize
            % (b) \imp
            (b) STS
        }}
        \put(79, 52){\parbox[t]{0.2\linewidth}{\footnotesize
            % (c) \diversity
            (c) GTD
        }}
        \put(11, 23){\parbox[t]{0.3\linewidth}{\footnotesize
            (d) Ours
        }}
        \put(50, 18){\parbox[t]{0.25\linewidth}{\footnotesize
            Salient objects
        }}
        \put(50, 7){\parbox[t]{0.2\linewidth}{\footnotesize
            Fine-grained structures
        }}
        \put(73, 1){\parbox[t]{0.3\linewidth}{\footnotesize
            Structural regions
        }}
    \end{overpic}
    % \vspace{0.5em} 
    \caption{\textbf{Qualitative ablation of token selection strategies.} (a) Full tokens without reduction.
    (b) Saliency-only selection (\imp) preserves salient objects but may miss fine-grained details like cables.
    (c) Diversity-only selection (\diversity) improves spatial coverage but can overlook semantically important regions on the tabletop.
    (d) Our method combines saliency and spatial diversity, retaining salient objects while preserving fine-grained structures and producing more continuous and complete structures for large objects such as the table.
    }
    \label{fig:ablation}
\end{figure}

\begin{table}[t]
\centering
\caption{Ablation Study of Components in Our Method.}
\resizebox{0.99\linewidth}{!}{
\label{tab:ablation1}
\begin{tabular}{l c c c}
    \toprule
    \textbf{Method} & \textbf{Strategy} & \textbf{ScanQA} & \textbf{OpenEQA} \\
    \midrule
    Base Model (Full) & None (100\% tokens) & 27.6 & 56.2 \\
    \midrule
    % \multicolumn{4}{l}{\textit{Compressed Models (Retention $\approx$ 25\%)}} \\
    \multicolumn{2}{l}{\textit{Compressed Models w/ Retention:}} & \textit{23\%} & \textit{26\%} \\
    \quad + Uniform & Uniform & 27.1 & 53.5 \\
    \quad + STS & Sal-only & 27.9 & 54.9 \\
    \quad + GTD & Div-only & 27.8 & 54.5 \\
    \quad \textbf{+ Ours} & \textbf{Sal + Div} & \textbf{28.0} & \textbf{55.0} \\
    \bottomrule
\end{tabular}
}
\end{table}

\subsection{Ablation Studies}
% \noindent{\textbf{Overall ablation.}}
\subsubsection{Overall ablation}

To further evaluate the effectiveness of each component in our proposed method, we conduct ablation studies on the \imp (STS) and \diversity(GTD) modules, with results summarized in Table~\ref{tab:ablation1}.
We conduct ablation studies at a representative retention ratio of approximately 25\%, where token reduction begins to significantly affect performance.
\textit{Uniform} denotes uniformly sampling visual tokens.
\textit{STS} corresponds to retaining only the top-$k$ tokens selected based on attention ranking.
\textit{GTD} represents sampling visual tokens solely based on the geometry-aware diversification strategy,

Uniform sampling yields poor performance on both datasets, as it neither preserves salient objects nor maintains sufficient spatial coverage, leading to severe information loss in complex 3D scenes.
On ScanQA~\cite{scanqa}, using either \imp or \diversity alone already improves performance over the base model.
When retaining 23\% of tokens, combining both components further yields a 0.4 EM@1 improvement over the base model.
On OpenEQA~\cite{openeqa}, both components significantly outperform uniform sampling, and their combination retains approximately 98\% of the base model performance under token reduction.
These results indicate that preserving salient semantics and maintaining spatial diversity are both essential for robust 3D reasoning under reduction.

To better understand the complementary roles of saliency preservation and spatial diversity, we further present qualitative ablation results in Fig.~\ref{fig:ablation}.
As illustrated in Fig.~\ref{fig:ablation}, \imp primarily preserves tokens corresponding to salient objects on the table, such as disinfectant wipes, but may miss fine-grained or peripheral structures.
In contrast, \diversity maintains broader spatial coverage by retaining tokens from different regions of the scene, including secondary objects and thin structures such as cables.
By integrating both components, \name not only preserves most object-level details on the table but also captures fine-grained elements across the scene.
Moreover, our method produces more coherent and complete structural representations for large objects, such as the table.
Overall, combining both components consistently achieves the strongest results across both datasets, demonstrating a clear synergistic effect.
These findings validate our design choice of jointly modeling saliency preservation and spatial diversity.

\begin{table}
    \centering
    % \caption{Comparison of Diverse Token Selection Strategies on the ScanQA}
    \caption{Comparison of Diversity-Aware Token Selection Strategies on ScanQA (Without Important Tokens).}
    \resizebox{0.98\linewidth}{!}{
    \begin{tabular}{l| c c c c c}
        \toprule
       \multirow{2}{*}{Method} & \multicolumn{5}{c}{Retention Ratio (\%)}\\
         &54 & 40 & 23 & 14& 9 \\
        \midrule
        Uniform &28.3 & 27.7& 27.1&\textbf{27.7} &21.2\\
        Semantic Similarity& 28.2& 28.0& 27.4& 27.0& 25.8 \\
        \textbf{GTD} & \textbf{28.4}&\textbf{28.3} & \textbf{27.8}& 26.7&\textbf{26.3}\\
        \bottomrule
    \end{tabular}
    \label{tab:sampling}
    }
\end{table}

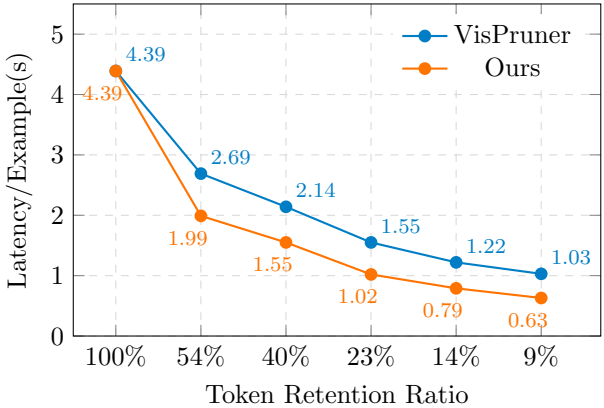
\begin{figure}[t]
    \centering
    \begin{tikzpicture}
    \begin{axis}[
        % --- 基础设置 ---
        width=0.98\linewidth,  % 图表宽度
        height=6cm,           % 图表高度
        grid=major,           % 显示网格
        grid style={dashed,gray!30}, % 网格样式
        % --- 坐标轴标签 ---
        xlabel={Token Retention Ratio},
        ylabel={Latency/Example(s)},
        ylabel near ticks,
        % ylabel style={yshift=-10pt},
        % --- X轴设置 (关键：处理非线性坐标) ---
        xtick={1,2,3,4,5,6}, % 指定X轴有6个刻度
        xticklabels={100\%, 54\%, 40\%, 23\%, 14\%, 9\%}, % 对应的标签
        xmin=0.5, xmax=6.7, % 控制X轴两端的留白
        % --- Y轴设置 ---
        ymin=0, ymax=5.5,     % Y轴范围，留出空间给上方的数字
        ytick={0,1,2,3,4,5},  % Y轴刻度
        % --- 图例设置 ---
        legend pos=north east, % 图例位置
        legend style={draw=none}, % 去掉图例边框
        % --- 数据点上的数字样式 ---
        nodes near coords, % 显示数值
        every node near coord/.append style={font=\footnotesize}, % 数字字体大小
    ]

    % --- 第一条线 (VisPruner - 蓝色) ---
    \addplot[
        color=cyan!60!blue, % 类似Excel的蓝色
        mark=*,             % 实心圆点
        thick,              % 线条加粗
        nodes near coords style={anchor=south west} % 数字位置微调(防重叠)
    ]
    coordinates {
        (1, 4.39) (2, 2.69) (3, 2.14) (4, 1.55) (5, 1.22) (6, 1.03)
    };
    \addlegendentry{VisPruner}

    % --- 第二条线 (Ours - 橙色) ---
    \addplot[
        color=orange!90!red, % 类似Excel的橙色
        mark=*,              % 实心圆点
        thick,
        nodes near coords style={anchor=north, xshift=-5pt, yshift=-2pt} % 数字放在点下方
    ]
    coordinates {
        (1, 4.39) (2, 1.99) (3, 1.55) (4, 1.02) (5, 0.79) (6, 0.63)
    };
    \addlegendentry{Ours}
        
    \end{axis}
    \end{tikzpicture}
    \caption{\textbf{Inference time comparison between VisPruner~\cite{vispruner} and ours} on the ScanQA~\cite{scanqa} validation set. Our method achieves lower latency per example at all token retention ratios. }
    \label{fig:speedup}
\end{figure}

% \noindent{\textbf{Other Methods for Diverse Tokens.}}
\subsubsection{Alternative Strategies for Diverse Token Selection}
To evaluate the effectiveness of our diverse token selection strategy, we compare several alternative sampling methods without using important tokens, as summarized in Table~\ref{tab:sampling}.
Specifically, we consider:
(1) Uniform Sampling, which uniformly samples visual tokens;
(2) Semantic Similarity Sampling, which selects tokens based solely on semantic similarity;
and (3) \diversity (GTD), which jointly considers semantic similarity and 3D spatial distance.

As shown in Table~\ref{tab:sampling}, Uniform Sampling performs poorly due to the absence of semantic and spatial guidance during token reduction, resulting in unstable and highly variable performance.
Although it achieves relatively strong results at a 14\% retention ratio, its performance drops sharply when the retention ratio is further reduced to 9\%.
Semantic Similarity Sampling, a commonly used strategy in 2D VLM token reduction~\cite{vispruner, sparsevlm}, demonstrates more stable performance than Uniform Sampling.

However, its performance consistently lags behind that of \diversity, highlighting the importance of incorporating spatial cues under severe token reduction.
By leveraging 3D spatial information, \diversity maintains stable and competitive performance across all retention ratios.
These results demonstrate that jointly modeling semantic relevance and spatial diversity is critical for effective token diversification and robust 3D reasoning.
This observation is consistent with our ablation results, further confirming the necessity of incorporating spatial information in the diverse token selection module.

% \input{table/speedup}

% \noindent{\textbf{Inference speedup.}}
\subsubsection{Inference speedup}
We evaluate the inference efficiency of VisPruner and \name under different token retention ratios, with the results summarized in Fig.~\ref{fig:speedup}.
As shown in the figure, both methods achieve notable inference speedups compared to the full-token baseline.
Nevertheless, \name consistently yields lower inference latency than VisPruner across all retention ratios, which demonstrates its ability to retain a compact yet informative token set by jointly modeling semantic relevance and 3D spatial diversity.
For instance, at a 9\% retention ratio, \name achieves a per-example latency of 0.63\,s, compared to 1.03\,s for VisPruner, corresponding to an approximate 39\% reduction in inference time.
Overall, these results demonstrate that incorporating explicit spatial cues into token reduction not only preserves 3D reasoning performance, as discussed in the previous sections, but also translates into tangible inference efficiency improvements.

\begin{figure*}[!htb]
    \centering
    \begin{overpic}[width=1.0\linewidth]{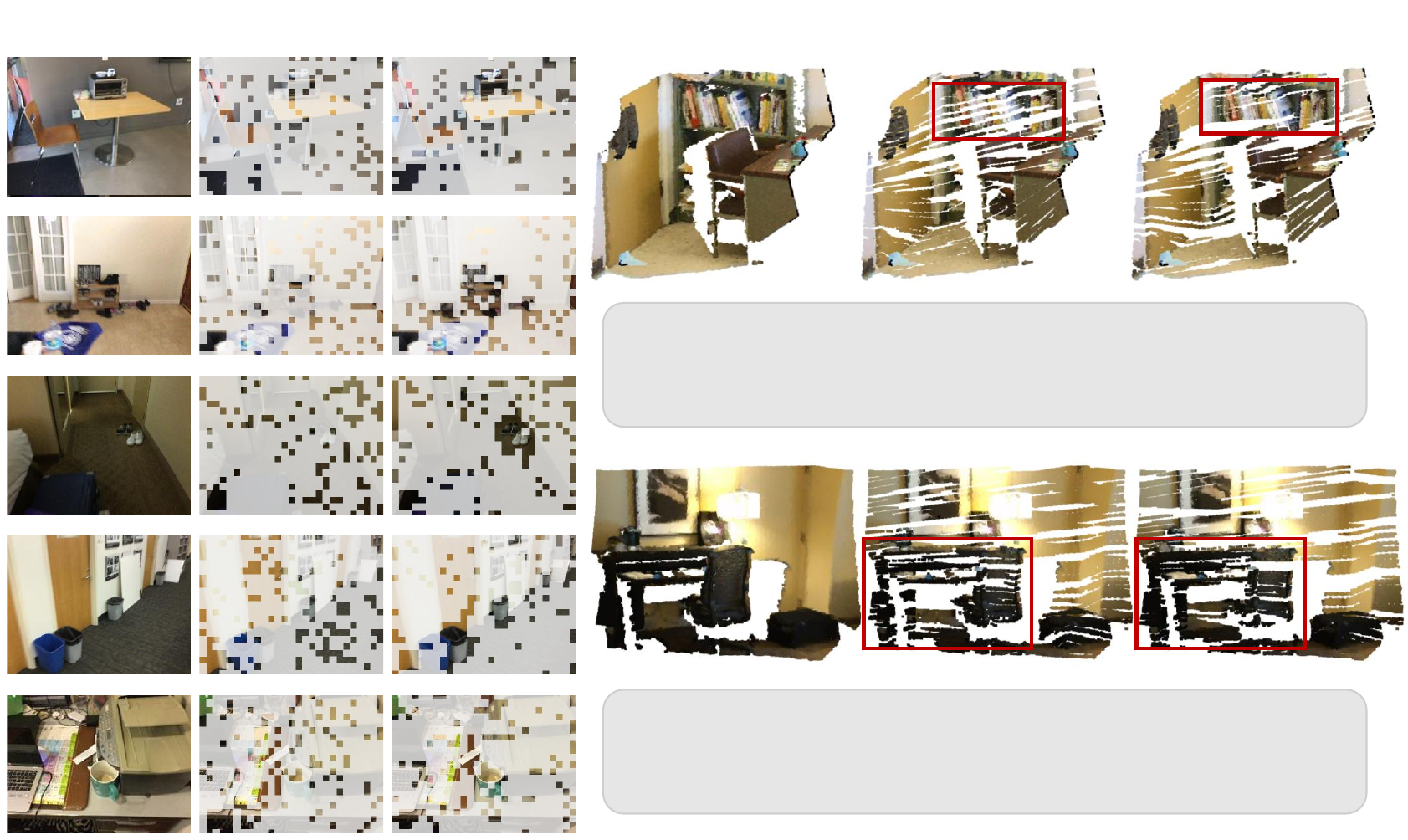}
        % \large
        \put(5, 56.5){\parbox[t]{0.2\linewidth}{\footnotesize
            Image
        }}
        \put(16, 56.5){\parbox[t]{0.2\linewidth}{\footnotesize
            VisPruner~\cite{vispruner}
        }}
        \put(33, 56.5){\parbox[t]{0.2\linewidth}{\footnotesize
            Ours
        }}
        \put(47, 56.5){\parbox[t]{0.3\linewidth}{\footnotesize
            Point Cloud
        }}
        \put(66, 56.5){\parbox[t]{0.25\linewidth}{\footnotesize
            VisPruner~\cite{vispruner}
        }}
        \put(88, 56.5){\parbox[t]{0.2\linewidth}{\footnotesize
            Ours
        }}
        \put(45, 35.5){\parbox[t]{0.5\linewidth}{\footnotesize
            \textcolor{blue}{Question}: Where is the door located?\\[2pt]
            VisPruner: to right of file cabinet\\[2pt]
            Ours: to right of bookshelf
        }}
        \put(69.5, 33){\textcolor{red}{\Large\ding{55}}}
        \put(65, 31){\textcolor{green!60!black}{\Large\ding{51}}} 
        
        \put(45, 8){\parbox[t]{0.5\linewidth}{\footnotesize
            \textcolor{blue}{Question}: What is underneath the painting?\\[2pt]
            VisPruner : toilet\\[2pt]
            Ours : desk
        }}
        \put(58, 5.5){\textcolor{red}{\Large\ding{55}}}
        \put(54, 3.3){\textcolor{green!60!black}{\Large\ding{51}}} 
        
    \end{overpic}
    % \vspace{0.5em} 
    \caption{\textbf{Qualitative comparison between our method and VisPruner on the ScanQA dataset.} \textit{Left:} retained tokens mapped onto the original image, with transparent areas denoting the selected tokens; our method achieves a more even spatial distribution of tokens in the image plane, thereby offering a more complete set of visual cues. 
    \textit{Right:} retained tokens projected into 3D space; compared with VisPruner, our approach produces a more continuous and complete point distribution, better capturing object structures with fewer holes and higher geometric completeness.
    }
    \label{fig:qualitative_study}
\end{figure*}
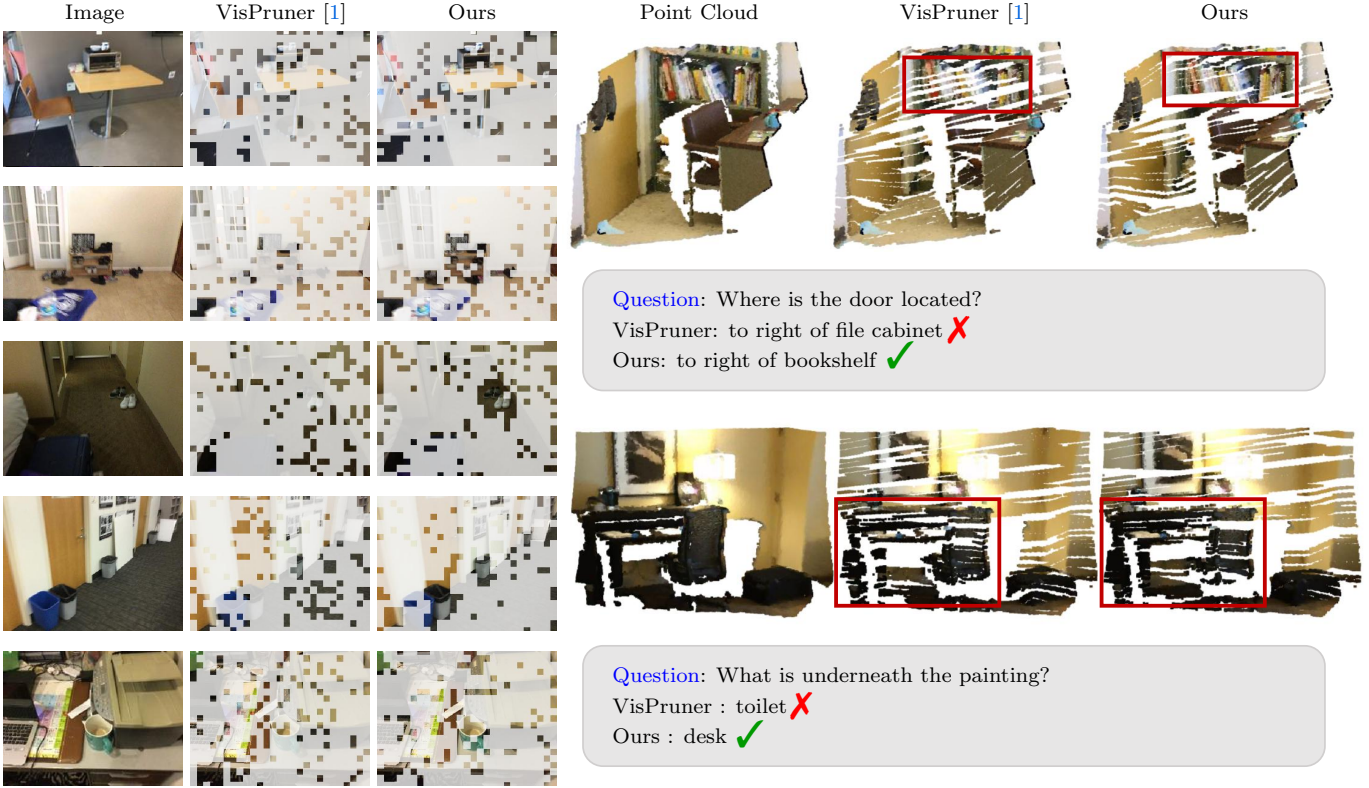

\subsection{Qualitative Results.}
We present qualitative visualizations on the ScanQA dataset in Fig.~\ref{fig:qualitative_study}.
Our method effectively retains key objects, while preserving the spatial distribution and semantic structure of the scene.
In contrast, VisPruner tends to discard secondary but critical objects or local details.
As illustrated in Fig.~\ref{fig:qualitative_study}, the left panels show retained tokens projected onto the image plane, while the right panels visualize the corresponding token distributions in 3D space.
In the image-plane visualizations, our method preserves a larger number of tokens on meaningful objects and fine-grained details, while avoiding excessive token allocation to large textureless regions such as walls.
In the 3D visualizations, our approach maintains more complete object structures, enabling clear identification of scene elements such as the books on the bookshelf in the first example.
These qualitative differences directly impact downstream reasoning.
As shown in the examples, our method enables more accurate object localization and spatial reasoning, leading to correct answers where VisPruner fails.
Overall, our approach preserves sufficient and reliable visual evidence for robust 3D question answering.

\section{Conclusion}
In this paper, we proposed \name, a training-free token reduction module for efficient multi-view 3D QA with off-the-shelf VLMs.
% \name integrates semantic relevance with 3D geometric cues to select a compact yet informative subset of multi-view visual tokens, suppressing redundancy while preserving object-critical evidence and spatially diverse context.
\name first identifies semantically relevant tokens to retain object-critical evidence for answering.
It then incorporates 3D geometric cues to select spatially diverse tokens, suppressing multi-view redundancy while maintaining broad contextual coverage.
Experiments on ScanQA and OpenEQA demonstrate that \name delivers a strong accuracy–efficiency trade-off, substantially reducing the number of visual tokens and accelerating inference while maintaining competitive performance.
Moreover, even under aggressive token budgets, \name preserves broad scene coverage, enabling robust 3D reasoning with a constrained inference budget.

% \clearpage
\bibliographystyle{IEEEtran}
\bibliography{ref}

\end{document}